\documentclass[11pt,a4paper]{article}
\usepackage[hyperref]{acl2021}
\usepackage{times}
\usepackage{latexsym}

\usepackage{amsmath}
\usepackage{amsfonts}
\usepackage{amsthm}
\usepackage{bbm}
\usepackage{booktabs}

\usepackage{algorithm}
\usepackage{algpseudocode}
\algnewcommand{\parState}[1]{\State%
  \parbox[t]{\dimexpr\linewidth-\algmargin}{\strut #1\strut}}
\algrenewcommand{\algorithmiccomment}[1]{\hskip1em$\rightarrow$ \footnotesize#1 \normalsize}

\usepackage{float}
\usepackage{multirow}

\usepackage{caption}
\usepackage{subcaption}

\usepackage{graphicx} 
\graphicspath{{./figs/}} 

\newcommand\citepossessive[1]{\citeauthor{#1}'s\ (\citeyear{#1})}

\usepackage{microtype}

\usepackage{cleveref}
\crefname{section}{\S}{\S\S}
\Crefname{section}{\S}{\S\S}
\crefname{table}{Tab.}{}
\crefname{figure}{Fig.}{}
\crefname{algorithm}{Algorithm}{}
\crefname{algorithm}{Algorithm}{}
\crefname{line}{Line}{}
\crefname{appendix}{App.}{}
\crefformat{section}{\S#2#1#3}
\crefname{thm}{Theorem}{}
\crefname{def}{Definition}{}
\crefname{prop}{Proposition}{}

\aclfinalcopy %
\def\aclpaperid{3004} %

\newcommand{\bz}{\mathbf{z}}
\newcommand{\bw}{{\boldsymbol w}}

\newcommand{\bl}{{\boldsymbol \ell}}

\newcommand{\vphi}{{\boldsymbol \phi}}

\newcommand{\pgen}{p_{\phi}}
\newcommand{\defn}[1]{\textbf{#1}}

\newcommand{\ent}{\mathrm{H}}

\newcommand{\word}[1]{\textit{#1}}

\newcommand{\LSTM}{\mathrm{LSTM}}
\newcommand{\pmodel}{p_{\mathrm{model}}}

\newcommand{\token}{\bw_n}

\newcommand{\cluster}{z_n}
\newcommand{\otherclusters}{\bz_{<n}}

\definecolor{modelred}{HTML}{FF0000}
\definecolor{modelgreen}{HTML}{008800}
\definecolor{modelblue}{HTML}{0000FF}
\definecolor{modelyellow}{HTML}{888800}

\newcommand{\twostagecolored}{{\color{modelred} \textbf{two-stage}}}
\newcommand{\tokencolored}{{\color{modelgreen} \textbf{token}}}
\newcommand{\typecolored}{{\color{modelblue} \textbf{type}}}
\newcommand{\generatorcolored}{{\color{modelyellow} \textbf{generator}}}

\everypar{\looseness=-1}
\title{Modeling the Unigram Distribution}

\usepackage{tipa}
\newcommand{\ucambridge}{\normalfont \text{\textipa{D}}}
\newcommand{\irenes}{\normalfont \text{\textipa{7}}}
\newcommand{\ethz}{\text{\normalfont \textipa{Q}}}
\newcommand{\harvard}{\normalfont \text{\textipa{@}}}

\newcommand{\mpi}{\normalfont \text{\textipa{9}}}
\newcommand{\hse}{\normalfont \text{\textipa{6}}}

\author{%
Irene Nikkarinen$\thanks{~~Equal contribution}$ $^{\,,\ucambridge,\irenes}$%
~\;~\;~ Tiago Pimentel$^{*,\ucambridge}$%
~\;~\;~ Dami\'{a}n E. Blasi$^{\harvard,\mpi,\hse}$
~\;~\;~ Ryan Cotterell$^{\ucambridge,\ethz}$ \\
  $^{\ucambridge}$University of Cambridge%
  ~\;~\;~\;~$^{\irenes}$Yle%
  ~\;~\;~\;~$^{\harvard}$Harvard University\\%
  $^{\mpi}$MPI for Evolutionary Anthropology%
  ~\;~\;~\;~$^{\hse}$HSE University%
  ~\;~\;~\;~$^{\ethz}$ETH Z\"{u}rich
   \\
  \texttt{irene.nikkarinen@gmail.com}%
  ,~\;~ \texttt{tp472@cam.ac.uk}\\%
  \texttt{dblasi@fas.harvard.edu}%
  ,~\;~ \texttt{ryan.cotterell@inf.ethz.ch}
 }

\date{}

\begin{document}
\maketitle
\begin{abstract}
The unigram distribution is the non-contextual probability of finding a specific word form in a corpus.
While of central importance to the study of language, it is commonly approximated by each word's sample frequency in the corpus.
This approach, being highly dependent on sample size, assigns zero probability to any out-of-vocabulary (oov) word form.
As a result, it produces negatively biased probabilities for any oov word form, while positively biased probabilities to in-corpus words. 
In this work, we argue in favor of properly modeling the unigram distribution---claiming it should be a central task in natural language processing.
With this in mind, we present a novel model for estimating it in a language (a neuralization of \citepossessive{Goldwater2011} model) and show it produces much better estimates across a diverse set of 7 languages than the na\"ive use of neural character-level language models.
\end{abstract}

\section{Introduction}

Neural networks have yielded impressive gains in sentence-level language modeling across a typologically diverse set of languages
 \cite{mikolov2010recurrent,kalchbrenner2016neural,merity2017regularizing,Melis2018,Cotterell2018}.
Similarly, neural networks constitute the state of the art in modeling
the distribution over a language's word types \cite{pimentel-etal-2020-phonotactics}, outperforming non-neural generative models such as \citepossessive{TACL971} with character-level models.
This paper focuses on a less-researched task that is halfway between sentence-level language modeling and word type distributions: Modeling the \defn{unigram distribution}, the distribution over word tokens in a language consisting of the probability of a word's form as well as its frequency in the language.
In particular, as opposed to sentence-level modeling, the unigram distribution does not consider contextual information.
\looseness=-1

The unigram distribution is a central object in the science of language from historical linguistics to psycholinguistics and beyond \cite{baayen2016frequency,diessel2017usage,divjak2019frequency}.
However, the majority of research on unigram distributions is based on identifying this distribution with sample frequency.
This approach results in poor estimates, as it assigns zero probability to out-of-vocabulary words.%
\footnote{In the Turkish Wikipedia, for example, 
considering a training set of $8$ million and a test set of $1$ million tokens, $27.4\%$ of test types and $5.3\%$ of test tokens are out-of-vocabulary.
Note that, according to Heaps' law, a similar behavior would be expected from corpora of any size \citep{herdan1960type,heaps1978information}.\looseness=-1} 
Further, it is highly dependent on sample size \cite{baayen2002word}

\begin{figure}
    \centering
    \includegraphics[width=\columnwidth]{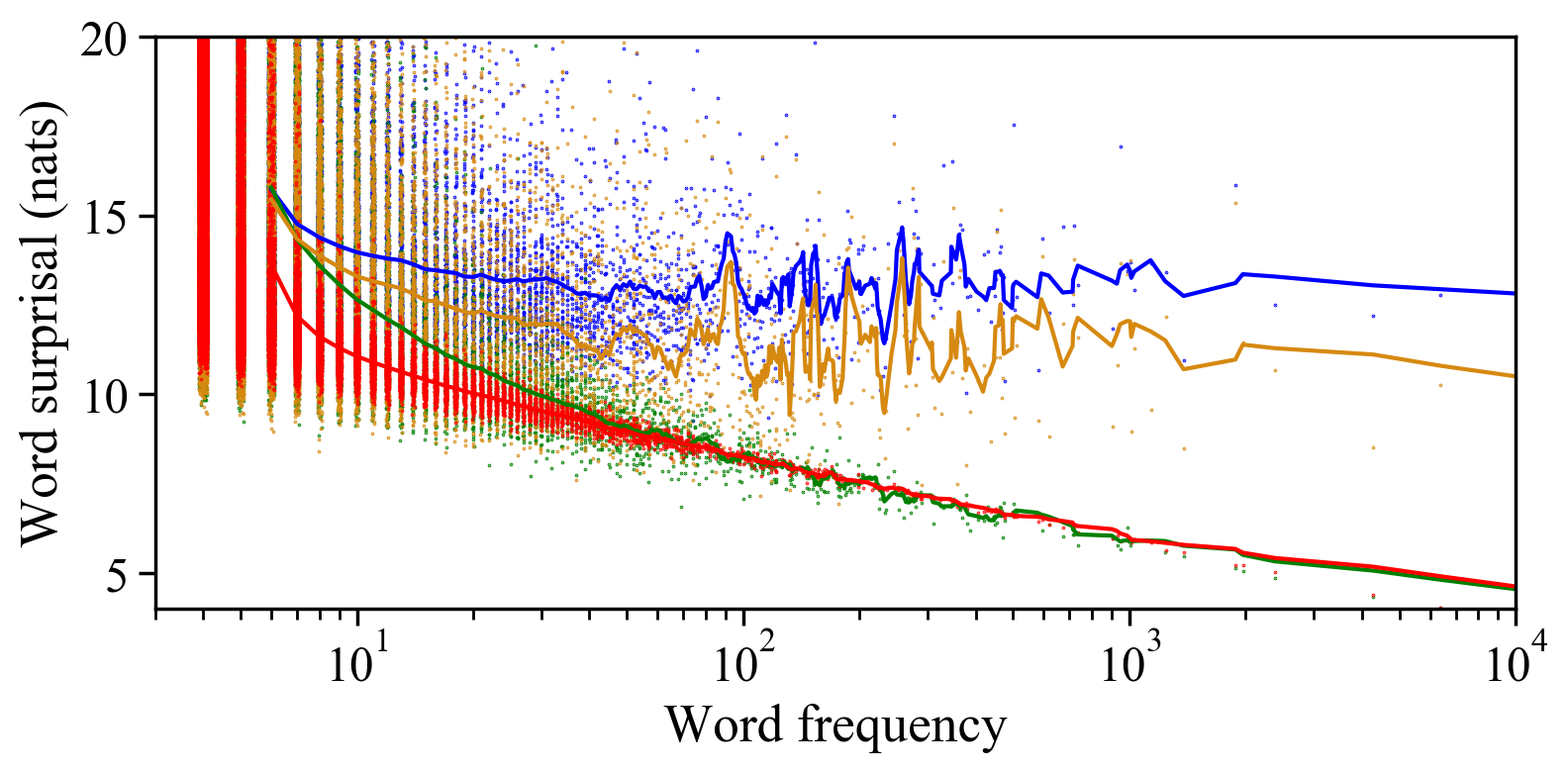}
    \caption{Word-level surprisal in Finnish under our \twostagecolored{} model, two baseline LSTMs trained with either word \typecolored{} or \tokencolored{} data, and another LSTM called the \generatorcolored{}, trained on an interpolation of both. Lines depict rolling averages.
    }
    \label{fig:average_entropy}
    \vspace*{-7pt}
\end{figure}

The core contribution of our work is motivating the unigram distribution as a worthwhile objective for scientific inquiry---one which is currently understudied in the field. 
With that in mind, we also present a neuralization of \citepossessive{Goldwater2011} two-stage model.\footnote{While \citet{Goldwater2011} acknowledge that their model could be used in various tasks of learning linguistic structure, they only present %
results in modeling morphology.\looseness=-1}
The gist of this approach is using two components to model the Zipfian distribution of word tokens in a language \cite{zipf1935psycho} separately from its phono- or graphotactic distribution. 
The first component, termed the \defn{adaptor}, 
is based on the Pitman--Yor process \citep[PYP;][]{Pitman1997}, and
has the ability to model the power-law behavior of word tokens. 
The second, termed the \defn{generator}, leverages a character-level neural language model to capture structural patterns in written words, e.g. graphotactics and morphology.

Critically, na\"{i}vely training a character-level neural model in either types (i.e. unique word forms) or tokens (i.e. word forms in their original frequencies) should lead to degenerate results. 
Models trained on natural corpora (i.e. token data) should excel in modeling the most common words of a language, but might poorly approximate the set of infrequent word forms which individuals dynamically produce (e.g. through compositional morphology).
On the other hand, training models on the collection of unique word forms (i.e. type data) would give equal weight to typical and atypical productions, potentially leading to poor performance on the most frequent forms, which any individual would recognize as part of their language.
In the two-stage model, as we will show, our generator is trained on a dataset interpolated between types and tokens---modeling the nuance between frequent and infrequent word forms better.

By testing our model on a set of languages with diverse morphological and phonological characteristics,
we find that it is capable of modeling both frequent and infrequent words, thus producing a better estimate of the unigram distribution than a character-level LSTM.
The empirical superiority of our two-stage model is shown in \cref{fig:average_entropy},
where the surprisal (i.e. the negative log-probability, measured in nats here) of each token is plotted under four different models for Finnish. 
Our proposed two-stage model
achieves a lower or similar surprisal to the baselines on tokens with all frequencies---with similar patterns arising in all analyzed languages.%
\footnote{As a final contribution of our work, the code used in this paper is available at \url{https://github.com/irenenikk/modelling-unigram}. We hope this will encourage future work in psycholinguistics to use the model to accurately investigate the effects of unigram probabilities in rare words.}

\section{The Unigram Distribution}

The unigram distribution is a probability distribution over the possible word forms in a language's lexicon.
This probability takes the frequency of a token into account, assigning larger probabilities to word forms which are more likely to be encountered in a language's utterances, thus differing from word type distributions, such as in \citet{pimentel-etal-2020-phonotactics}.
It is also not conditioned on a word's context, as it considers each word token as a stand-alone unit, as opposed to the task of language modeling, e.g. \citet{mikolov2010recurrent}.%

\subsection{Complex Vocabularies}

The composition of spoken vocabularies is structured according to a host of factors.
Stemming from articulatory biases, each language has a set of constraints on what sequences of speech sounds can be valid words in it; this is termed the \defn{phonotactics} of a language.
Languages also exhibit small but non-negligible biases in the regular match of forms and meanings  \cite{dingemanse2015arbitrariness,pimentel2019meaning,pimentel-etal-2021-finding}.
Additionally, expectations about morphology can constrain the production or processing of a given word as belonging to a particular word class (as shown for instance in Jabberwocky- and wug-type tasks, \citealt{berko1958child}, \citealt{hall-maudslay-cotterell-2021-syntactic}).
\looseness=-1

While individuals often have strong intuitions about these patterns, their judgments are typically gradient rather than categorical \cite{hayes2008maximum,gorman2013generative}.
The effective set of words that naturally occur in linguistic productions are known to be extremely diverse in their composition.
Models deployed to explain and predict typical word forms in a given language might fail at capturing these %
corners of the space of possible forms.
If the goal is to produce ecologically valid models that could approximate  actual cognitive processes, these atypical forms should be efficiently learned in addition to the most typical productions.

\subsection{Imbalanced Frequencies}

\newcite{zipf1935psycho} popularized the observation that the frequency of a word in a corpus is inversely proportional to its rank, approximately following a power-law distribution.
As such, a small subset of the most common word types dominate the corpus.
These extremely frequent words tend to be short in length and exceptionally archaic, in the sense that they preserve traces of previous phonotactic and phonological profiles that might have ceased to be productive.
This is particularly relevant when we consider scenarios where substantial portions of the vocabulary might have been borrowed from different sources over time. 
English is a textbook example: \citet{williams1986origins} reports that French, Latin, Germanic and Greek account for 29\%, 29\%, 26\% and 6\% of all words' origins in the vocabulary (plus a remaining 10\% of diverse origin). 
The most frequent portion of the vocabulary preserves the most the original West Germanic forms, consisting largely of articles, prepositions, pronouns, and auxiliaries.
Further, irregular inflections tend to be more common in these highly frequent words \cite{Ackerman2013,Cotterell2019}.
This observation might invite one to omit frequency information from training data, i.e. to use types, in order to balance out the role of the most frequent words.

On the other side of the frequency scale, however, any natural language data would have plenty of low-frequency words that reflect the open boundaries of the vocabulary. These might include nonce words (\word{blick}), expressive transformations of other words (\word{a loooooooooooong summer}), specialized terms (\word{onabotulinumtoxina}), and names, among others. In addition, genuine orthographic misproductions (\word{langague}) will be present to some degree.

Finally, acronyms (\word{HTML}) will be present in all frequency bands.
These should be particularly problematic to model, since they do not necessarily follow the language's graphotactics to any degree.
There are also frequent and infrequent loanwords with different degrees of adjustment to the grapho- and phonotactics of the rest of the vocabulary. 
For instance, it has been estimated that 96\% and 21\% of English speakers know the Afrikaans-originated words \word{aardvark} and \word{aardwolf}, respectively \citep{brysbaert2019word}.\footnote{Aardvarks and aardwolves are African mammals.}
These are the only written word forms in English with a non-negligible frequency that display two letter `\word{a}'s in word-initial position.
\looseness=-1

This whimsical nature of the vocabulary of a language makes modeling the unigram distribution challenging:
Na{\"i}vely training a model to capture word forms at either the token or type level is likely to give disproportionate emphasis to phonotactically unrepresentative words.
However, this is also why its modeling is a worthwhile task---it captures
both frequent and rare productions, combining form probability with frequency information.

\section{Modeling the Unigram Distribution}

Our work neuralizes \citepossessive{Goldwater2011} two-stage model and employs it to modeling the unigram distribution.\footnote{This same model was used in our contemporary work investigating lexicons' (non-)optimality \citep{pimentel-etal-2021-hownonoptimal}.}
The first component, termed the \textbf{generator}, is a model used to produce a set of i.i.d. word forms $\{\bl_k\}_{k=1}^{K}$.
The second component is termed \textbf{adaptor}, and it assigns each instance in the  training set to a cluster $\{z_n\}_{n=1}^{N}$.
Under this model, each token in a dataset has a corresponding cluster $z_n$ which defines the token's word form $\bw_n=\bl_{z_n}$.
We note that both word forms $\bl$ and clusters $\bz$ are latent variables, and only tokens $\bw$ are observed during training.

\paragraph{Generator.}
The generator is a model which produces word forms;
we use a character-level LSTM here \citep{hochreiter1997long}, as in:\footnote{See \citet{pimentel-etal-2020-phonotactics} for more details on this graphotactics generative model.}
\begin{align}
    & \{\bl_k\}_{k=1}^{K} \sim \pgen(\bl) = \LSTM(\bl)
\end{align}
These word forms $\bl_k$ are sampled i.i.d.---thus, the same word may be sampled more than once.

\paragraph{Adaptor.}
Each word form sampled from the generator corresponds to a cluster.
The adaptor then assigns a frequency to each of the clusters according to a Pitman--Yor process:
\begin{align}
    p&(\cluster \mid\, \otherclusters) \\
    &\propto\begin{cases}
        c_{<n}^{(\cluster)} - a  & 1 \leq \cluster \leq K_{<n}~~ \text{\color{gray} (old\ cluster)} \\
        a \cdot K_{<n} + b \, & \cluster = K_{<n} + 1~ \text{\color{gray} (new\ cluster)} \nonumber
    \end{cases} 
    \label{eq:table_assignment}
\end{align}
where $0 \leq a < 1$ and $0 \leq b$ are hyperparameters of the PYP, $\otherclusters$ are the previous cluster assignments,
$K_{<n}$ is the current number of clusters with at least one token and $c_{<n}^{(\cluster)}$ is the number of tokens previously assigned to cluster $\cluster$.
This adaptor, as a Pitman--Yor process, allows us to model the power-law distribution of word tokens.

\paragraph{Two-stage Model.}
Given a cluster assignment and the list of word forms, defining a token's form is deterministic: 
$p(\bw_n \mid z_n, \bl) = \mathbbm{1}\{\bw_n = \bl_{z_n}\}$.
Thus, our model factorizes a new token's probability into two terms:
\begin{align} 
    &\pmodel(\bw) = \\
    &\quad \underbrace{\frac{ c_\bw - \overbrace{n_\bw \cdot a}^{\text{smoothing factor}}}{|\bz| + b}}_{\text{smoothed $1$-gram}} + \underbrace{\frac{(a \cdot K + b)}{|\bz| + b}}_{\text{interpolation weight}} \cdot \underbrace{\pgen(\bw)}_{\text{LSTM}} \nonumber
\end{align}
where $c_\bw$ is the number of occurrences of word form $\bw$ in our training corpus and $n_\bw$ is the number of distinct clusters to which it has been assigned:
\begin{align}
    c_{\bw} &= \sum_{n=1}^{N} \mathbbm{1}\{\bw = \bl_{\cluster}\}, \\
    n_{\bw} &= \sum_{k=1}^{K} \mathbbm{1}\{\bw  = \bl_k\} 
\end{align}
In practice, the two-stage model acts as an interpolation between a smoothed $1$-gram model, i.e. corpus frequencies, and an LSTM character model. 
Notably, this model learns per-word smoothing factors and its interpolation weight in an unsupervised manner through the PYP parameters' inference.
The adaptor is fit using Gibbs sampling, and the generator is trained using a cross-entropy loss on the set of non-empty clusters produced by the adaptor. The generator is thus trained using a more balanced corpus where the proportion of the most frequent words is reduced; this can be seen as an interpolation between a type and a token dataset.%
\footnote{This model's training is detailed in \cref{sec:inference}. For a detailed description of the adaptor see \citet{Goldwater2011}.\looseness=-1}

\section{Experiments.}

\paragraph{Dataset.}
We use Wikipedia data and evaluate our model on the following languages: English, Finnish, Hebrew, Indonesian, Tamil, Turkish and Yoruba. These languages represent a typologically diverse set---with different levels of morphology, ranging from rich (e.g. Finnish) to poor (e.g. Yoruba), as well as distinct scripts and graphotactic patterns. 
In preprocessing, we first split the data into sentences and then into tokens using spaCy \cite{spacy}.
We then sample $10^6$ tokens as our training set for each language (except for Yoruba for which we had less data, see \cref{sec:experimental_setup} for more details).
From these, we build two distinct datasets: a \defn{token dataset}, which corresponds to the list of word forms with their corpus frequency, and a \defn{type dataset} containing the set of unique word forms in the data.

\paragraph{Evaluation.}
We measure the cross-entropy of our models on a
held-out test set; this is the standard evaluation for language modeling.
We approximate this cross-entropy using a sample mean estimate
\begin{align}
    \ent(p) &\leq \ent(p, \pmodel) \\
    &\approx - \frac{1}{N} \sum_{n=1}^N \log \pmodel(\token) \nonumber
\end{align}
where we assume instances $\token$ are sampled from the true unigram distribution $p(\bw)$.
Specifically, these token samples $\{\token\}_{n=1}^{N}$ take the form of the token dataset.
The model with the lowest cross-entropy is the one that diverges the least from the true distribution.

\paragraph{Baseline Models.}
As neural networks yield state-of-the-art performance in language modeling tasks, we expect them to also do well with the unigram distribution.
In fact, pseudo-text generated by LSTM-based language models reproduces Zipf's law to some extent \citep{takahashi2017neural,meister-etal-2021-language}.
Thus, we view state-of-the-art LSTM models as a strong baseline.
We train a character-level LSTM language model \citep{pimentel-etal-2020-phonotactics} to directly approximate the unigram distribution by training it on the token dataset---modeling these tokens at the character level.
As a second baseline, we train an LSTM on the type dataset.
However, we expect this model to be outperformed by the token one in the unigram distribution task, as the information on word frequency is not available during its training.
We do not use a word-level $1$-gram model (i.e. the words' sample frequency) as a baseline here, since it results in an infinite cross-entropy for any test set containing oov words.
We empirically compare four models: \defn{two-stage}, \defn{generator}, \defn{token}, and \defn{type}.\looseness=-1

\begin{table}[t]
    \centering
\resizebox{\columnwidth}{!}{%
    \begin{tabular}{lcccc} \toprule
        & \multicolumn{4}{c}{Evaluated Model} \\ \cmidrule(lr){2-5}
        Language & Type & Token & Two-stage & Generator \\
        \midrule
        English & 12.50 & \phantom{0}9.04 & \phantom{0}\textbf{8.34} & 11.86 \\
        Finnish & 15.07 & 12.94 & \textbf{11.85} & 14.16 \\
        Hebrew & 12.62 & 10.71 & \textbf{10.20} & 11.76 \\
        Indonesian & 13.28 & 10.33 & \phantom{0}\textbf{9.55} & 11.89 \\
        Tamil & 14.24 & 12.66 & \textbf{11.76} & 13.29 \\
        Turkish & 14.00 & 11.77 & \textbf{10.86} & 12.95 \\
        Yoruba & 11.13 & 10.04 & \phantom{0}\textbf{9.19} & \phantom{0}9.88\\  \bottomrule
    \end{tabular}
}
    \caption{
    Cross-entropy on the \textbf{unigram distribution}.
    }
    \label{tab:token_test_cross_entropies}
\end{table}

\paragraph{Modeling Tokens.}
Cross-entropy on the token test sets can be found in \cref{tab:token_test_cross_entropies}.
These results show our two-stage model indeed creates a more accurate estimate of the unigram distribution, producing the smallest cross-entropy across all languages.

\paragraph{Frequent vs Infrequent Words.}
The weaknesses of the token and type models are evinced by \cref{fig:average_entropy}.
In line with our hypothesis, the token model achieves lower cross-entropy on the most common words, but fails to model the rare ones accurately.
The cross-entropy achieved by the type model does not change as much with word frequency, but is higher than the one achieved by the token model for most of the vocabulary.
We also see that the two-stage model performs well across all word frequencies.
Indeed, this model appears to behave similarly to the token model with frequent words, but obtains a lower cross-entropy on the rare ones, where the role of the generator in the estimated probability is emphasized.
We suspect this is the reason behind the two-stage model's success.

\begin{table}[t]
    \centering
\resizebox{\columnwidth}{!}{%
    \begin{tabular}{lcccc|c} \toprule
        & \multicolumn{4}{c}{Evaluated Model} \\ \cmidrule(lr){2-5}
        Language & Type & Token & Two-stage & Generator & \% \\
        \midrule
        English & \textbf{18.23} & 21.71 & 20.32 & 18.59 & 56\% \\
        Finnish & \textbf{19.76} & 21.42 & 19.79 & 19.89 & 71\% \\
        Hebrew & \textbf{15.81} & 17.76 & 17.10 & 16.30 & 56\% \\
        Indonesian & \textbf{18.52} & 21.20 & 19.15 & 19.17 & 61\% \\
        Tamil & \textbf{18.77} & 20.37 & 19.26 & 19.19 & 71\% \\
        Turkish & \textbf{18.36} & 19.78 & 18.50 & 18.62 & 65\% \\
        Yoruba & 15.44 & 17.69 & \textbf{15.34} & 16.28 & 67\% \\  \bottomrule
    \end{tabular}
}
    \caption{Average surprisal for \textbf{singleton} types. Column \% represents the ratio of singletons in the type test set.}
    \label{tab:singletons}
\end{table}

\paragraph{The Long Tail.}
\Cref{fig:average_entropy} also demonstrates that the entropy estimate for the rare words grows quickly and exhibits a large variance across models.
This reflects the heterogeneous nature of the words that only appear a few times in a corpus.
This part of the vocabulary is where the type model achieves the best results for all languages except Yoruba (see \cref{tab:singletons}).\footnote{We note that we used considerably less training data for Yoruba than for other languages.}
The fact that singletons (also known as \textit{hapax legomena}), i.e. word forms which occur only once in the test set, form a large portion of the type dataset boosts the type model's performance on rare words.
However, in the case of words appearing more than once (see \cref{tab:non_singletons}) the two-stage model achieves the best results across languages.
Furthermore, in these non-singleton words, the generator outperforms the type and token models in all languages except for Yoruba.
This shows the utility of training the generator on an interpolation between types and tokens. 
In addition, we note that one may justifiably question whether properly modeling singletons is a desirable feature, since they are likely to contain unrepresentative word forms, such as typos, as discussed previously.
Indeed, it appears that the two-stage model not only leads to tighter estimates of the unigram distribution, but also allows us to train a better graphotactics model; capable of modeling both frequent word forms as well as new productions.

\begin{table}[t]
    \centering
\resizebox{\columnwidth}{!}{%
    \begin{tabular}{lcccc} \toprule
        & \multicolumn{4}{c}{Evaluated Model} \\ \cmidrule(lr){2-5}
        Language & Type & Token & Two-stage & Generator\\
        \midrule
        English & 14.50 & 15.94 & \textbf{13.24} & 14.25 \\
        Finnish & 15.19 & 14.89 & \textbf{12.73} & 14.80\\
        Hebrew & 13.36 & 13.80 & \textbf{12.66} & 13.20 \\
        Indonesian & 14.72 & 15.50 & \textbf{12.80} & 14.64 \\
        Tamil & 14.65 & 14.77 & \textbf{12.95} & 14.33 \\
        Turkish & 14.73 & 14.41 & \textbf{12.41} & 14.33 \\
        Yoruba & 11.13 & 11.97 & \textbf{10.36} & 11.16 \\  \bottomrule
    \end{tabular}
}
    \caption{The average surprisal for \textbf{non-singleton} types.
    }
    \label{tab:non_singletons}
\end{table}

\paragraph{Future Work.}
The results we present focus on analyzing the two-stage model. The generator, though, produces interesting results by itself, modeling non-singleton word forms better than the type and token models in most languages. This suggests that it might be better at modeling the graphotactics of a language than either of these baselines.
Future work should explore if this indeed is the case. 

\section{Conclusion}

In this work, we motivate the unigram distribution as an important task to both the psycholinguistics and natural language processing communities that has received too little attention.
We present a two-stage model for estimating this distribution---a neuralization of \citepossessive{Goldwater2011}---%
which is motivated by the complex makeup of vocabularies:
This model defines the probability of a token by combining the probability of its appearance in the training corpus with the probability of its form.
We have shown, through a cross-entropy evaluation, that our model outperforms na\"ive solutions and is capable of accurately modeling both frequent and infrequent words.

\section*{Acknowledgements}

Dami\'{a}n E. Blasi acknowledges funding from the Branco Weiss Fellowship, administered by the ETH Z\"{u}rich.
Dami\'{a}n E. Blasi's research was also executed within the framework of the HSE University Basic Research Program and funded by the Russian Academic Excellence Project `5-100'.

\section*{Ethical Concerns}

This paper highlights the importance of modeling the unigram distribution, and presents a model for the task. We do not foresee any reasons for ethical concerns, but we would like to note that the use of Wikipedia as a data source may introduce some bias into our experiments.

\bibliography{acl2020}

\begin{thebibliography}{36}
\expandafter\ifx\csname natexlab\endcsname\relax\def\natexlab#1{#1}\fi

\bibitem[{Ackerman and Malouf(2013)}]{Ackerman2013}
Farrell Ackerman and Robert Malouf. 2013.
\newblock \href {https://muse.jhu.edu/article/521667} {Morphological
  organization: {T}he low conditional entropy conjecture}.
\newblock \emph{Language}, 89(3):429--464.

\bibitem[{Baayen(2002)}]{baayen2002word}
R.~Harald Baayen. 2002.
\newblock \href {https://www.springer.com/gp/book/9780792370178} {\emph{Word
  Frequency Distributions}}, volume~18.
\newblock Springer Science \& Business Media.

\bibitem[{Baayen et~al.(2016)Baayen, Milin, and Ramscar}]{baayen2016frequency}
R.~Harald Baayen, Petar Milin, and Michael Ramscar. 2016.
\newblock \href {https://doi.org/10.1080/02687038.2016.1147767} {Frequency in
  lexical processing}.
\newblock \emph{Aphasiology}, 30(11):1174--1220.

\bibitem[{Bergstra and Bengio(2012)}]{Bergstra2012}
James Bergstra and Yoshua Bengio. 2012.
\newblock \href
  {https://www.jmlr.org/papers/volume13/bergstra12a/bergstra12a.pdf} {Random
  search for hyper-parameter optimization}.
\newblock \emph{Journal of Machine Learning Research}, 13:281--305.

\bibitem[{Berko(1958)}]{berko1958child}
Jean Berko. 1958.
\newblock \href {https://doi.org/10.1080/00437956.1958.11659661} {The child's
  learning of {E}nglish morphology}.
\newblock \emph{Word}, 14(2-3):150--177.

\bibitem[{Blunsom et~al.(2009)Blunsom, Cohn, Goldwater, and
  Johnson}]{blunsom-etal-2009-note}
Phil Blunsom, Trevor Cohn, Sharon Goldwater, and Mark Johnson. 2009.
\newblock \href {https://www.aclweb.org/anthology/P09-2085} {A note on the
  implementation of hierarchical {D}irichlet processes}.
\newblock In \emph{Proceedings of the {ACL}-{IJCNLP} 2009 Conference Short
  Papers}, pages 337--340, Suntec, Singapore. Association for Computational
  Linguistics.

\bibitem[{Brysbaert et~al.(2019)Brysbaert, Mandera, McCormick, and
  Keuleers}]{brysbaert2019word}
Marc Brysbaert, Pawe{\l} Mandera, Samantha~F. McCormick, and Emmanuel Keuleers.
  2019.
\newblock \href {https://link.springer.com/article/10.3758/s13428-018-1077-9}
  {Word prevalence norms for 62,000 {E}nglish lemmas}.
\newblock \emph{Behavior Research Methods}, 51(2):467--479.

\bibitem[{Cotterell et~al.(2019)Cotterell, Kirov, Hulden, and
  Eisner}]{Cotterell2019}
Ryan Cotterell, Christo Kirov, Mans Hulden, and Jason Eisner. 2019.
\newblock \href {http://arxiv.org/abs/1807.02747} {On the complexity and
  typology of inflectional morphological systems}.
\newblock \emph{Transactions of the Association for Computational Linguistics},
  7:327--342.

\bibitem[{Cotterell et~al.(2018)Cotterell, Mielke, Eisner, and
  Roark}]{Cotterell2018}
Ryan Cotterell, Sabrina~J. Mielke, Jason Eisner, and Brian Roark. 2018.
\newblock \href {https://doi.org/10.18653/v1/N18-2085} {Are all languages
  equally hard to language-model?}
\newblock In \emph{Proceedings of the 2018 Conference of the North {A}merican
  Chapter of the Association for Computational Linguistics: Human Language
  Technologies, Volume 2 (Short Papers)}, pages 536--541, New Orleans,
  Louisiana. Association for Computational Linguistics.

\bibitem[{Diessel(2017)}]{diessel2017usage}
Holger Diessel. 2017.
\newblock \href {https://doi.org/10.1093/acrefore/9780199384655.013.363}
  {Usage-based linguistics}.
\newblock In \emph{Oxford Research Encyclopedia of Linguistics}. Oxford
  University Press.

\bibitem[{Dingemanse et~al.(2015)Dingemanse, Blasi, Lupyan, Christiansen, and
  Monaghan}]{dingemanse2015arbitrariness}
Mark Dingemanse, Dami{\'a}n~E. Blasi, Gary Lupyan, Morten~H. Christiansen, and
  Padraic Monaghan. 2015.
\newblock \href
  {https://www.sciencedirect.com/science/article/pii/S1364661315001771}
  {Arbitrariness, iconicity, and systematicity in language}.
\newblock \emph{Trends in Cognitive Sciences}, 19(10):603--615.

\bibitem[{Divjak(2019)}]{divjak2019frequency}
Dagmar Divjak. 2019.
\newblock \href {https://doi.org/10.1017/9781316084410} {\emph{Frequency in
  Language: {M}emory, Attention and Learning}}.
\newblock Cambridge University Press.

\bibitem[{Futrell et~al.(2017)Futrell, Albright, Graff, and
  O'Donnell}]{TACL971}
Richard Futrell, Adam Albright, Peter Graff, and Timothy O'Donnell. 2017.
\newblock \href {https://transacl.org/ojs/index.php/tacl/article/view/971} {A
  generative model of phonotactics}.
\newblock \emph{Transactions of the Association for Computational Linguistics},
  5(0):73--86.

\bibitem[{Goldwater et~al.(2011)Goldwater, Griffiths, and
  Johnson}]{Goldwater2011}
Sharon Goldwater, Thomas~L. Griffiths, and Mark Johnson. 2011.
\newblock \href {http://jmlr.org/papers/v12/goldwater11a.html} {Producing
  power-law distributions and damping word frequencies with two-stage language
  models}.
\newblock \emph{Journal of Machine Learning Research}, 12(68):2335--2382.

\bibitem[{Gorman(2013)}]{gorman2013generative}
Kyle Gorman. 2013.
\newblock \href {https://repository.upenn.edu/dissertations/AAI3565062}
  {\emph{Generative Phonotactics}}.
\newblock University of Pennsylvania.

\bibitem[{Hall~Maudslay and
  Cotterell(2021)}]{hall-maudslay-cotterell-2021-syntactic}
Rowan Hall~Maudslay and Ryan Cotterell. 2021.
\newblock \href {https://www.aclweb.org/anthology/2021.naacl-main.11} {Do
  syntactic probes probe syntax? experiments with jabberwocky probing}.
\newblock In \emph{Proceedings of the 2021 Conference of the North American
  Chapter of the Association for Computational Linguistics: Human Language
  Technologies}, pages 124--131, Online. Association for Computational
  Linguistics.

\bibitem[{Hayes and Wilson(2008)}]{hayes2008maximum}
Bruce Hayes and Colin Wilson. 2008.
\newblock \href {https://doi.org/10.1162/ling.2008.39.3.379} {A maximum entropy
  model of phonotactics and phonotactic learning}.
\newblock \emph{Linguistic Inquiry}, 39(3):379--440.

\bibitem[{Heaps(1978)}]{heaps1978information}
Harold~Stanley Heaps. 1978.
\newblock \href {https://dl.acm.org/doi/book/10.5555/539986} {\emph{Information
  Retrieval, Computational and Theoretical Aspects}}.
\newblock Academic Press.

\bibitem[{Herdan(1960)}]{herdan1960type}
Gustav Herdan. 1960.
\newblock \href
  {https://www.amazon.com/Type-token-mathematics-textbook-mathematical-linguistics/dp/B000PKJREU}
  {\emph{Type--Token Mathematics: {A} Textbook of Mathematical Linguistics}},
  volume~4.
\newblock Mouton.

\bibitem[{Hochreiter and Schmidhuber(1997)}]{hochreiter1997long}
Sepp Hochreiter and J{\"u}rgen Schmidhuber. 1997.
\newblock \href {https://doi.org/10.1162/neco.1997.9.8.1735} {Long short-term
  memory}.
\newblock \emph{Neural Computation}, 9(8):1735--1780.

\bibitem[{Honnibal et~al.(2020)Honnibal, Montani, Van~Landeghem, and
  Boyd}]{spacy}
Matthew Honnibal, Ines Montani, Sofie Van~Landeghem, and Adriane Boyd. 2020.
\newblock \href {https://doi.org/10.5281/zenodo.1212303} {{spaCy}:
  {I}ndustrial-strength natural language processing in python}.

\bibitem[{Kalchbrenner et~al.(2016)Kalchbrenner, Espeholt, Simonyan, van~den
  Oord, Graves, and Kavukcuoglu}]{kalchbrenner2016neural}
Nal Kalchbrenner, Lasse Espeholt, Karen Simonyan, Aäron van~den Oord,
  Alexander Graves, and Koray Kavukcuoglu. 2016.
\newblock \href {https://arxiv.org/abs/1610.10099} {Neural machine translation
  in linear time}.
\newblock In \emph{arXiv preprint arXiv:1610.10099}.

\bibitem[{Meister and Cotterell(2021)}]{meister-etal-2021-language}
Clara Meister and Ryan Cotterell. 2021.
\newblock \href {https://arxiv.org/abs/2106.00085} {Language model evaluation
  beyond perplexity}.
\newblock In \emph{Proceedings of the 59th Annual Meeting of the Association
  for Computational Linguistics}, Online. Association for Computational
  Linguistics.

\bibitem[{Melis et~al.(2018)Melis, Dyer, and Blunsom}]{Melis2018}
G{\'a}bor Melis, Chris Dyer, and Phil Blunsom. 2018.
\newblock \href {https://openreview.net/forum?id=ByJHuTgA-} {On the state of
  the art of evaluation in neural language models}.
\newblock In \emph{International Conference on Learning Representations}.

\bibitem[{Merity et~al.(2018)Merity, Keskar, and
  Socher}]{merity2017regularizing}
Stephen Merity, Nitish~Shirish Keskar, and Richard Socher. 2018.
\newblock \href {https://openreview.net/forum?id=SyyGPP0TZ} {Regularizing and
  optimizing {LSTM} language models}.
\newblock In \emph{International Conference on Learning Representations}.

\bibitem[{Mikolov et~al.(2010)Mikolov, Karafi{\'a}t, Burget,
  {\v{C}}ernock{\'y}, and Khudanpur}]{mikolov2010recurrent}
Tom{\'a}{\v{s}} Mikolov, Martin Karafi{\'a}t, Luk{\'a}{\v{s}} Burget, Jan
  {\v{C}}ernock{\'y}, and Sanjeev Khudanpur. 2010.
\newblock \href
  {https://www.isca-speech.org/archive/archive_papers/interspeech_2010/i10_1045}
  {Recurrent neural network based language model}.
\newblock In \emph{Eleventh annual conference of the International Speech
  Communication Association}, pages 1045--1048.

\bibitem[{Neal(1993)}]{Neal1993}
Radford~M. Neal. 1993.
\newblock \href {https://www.cs.toronto.edu/~radford/ftp/review.pdf}
  {Probabilistic inference using {M}arkov chain {M}onte {C}arlo methods}.
\newblock Technical report, University of Toronto.

\bibitem[{Pimentel et~al.(2019)Pimentel, McCarthy, Blasi, Roark, and
  Cotterell}]{pimentel2019meaning}
Tiago Pimentel, Arya~D. McCarthy, Damian Blasi, Brian Roark, and Ryan
  Cotterell. 2019.
\newblock \href {https://doi.org/10.18653/v1/P19-1171} {Meaning to form:
  Measuring systematicity as information}.
\newblock In \emph{Proceedings of the 57th Annual Meeting of the Association
  for Computational Linguistics}, pages 1751--1764, Florence, Italy.
  Association for Computational Linguistics.

\bibitem[{Pimentel et~al.(2021{\natexlab{a}})Pimentel, Nikkarinen, Mahowald,
  Cotterell, and Blasi}]{pimentel-etal-2021-hownonoptimal}
Tiago Pimentel, Irene Nikkarinen, Kyle Mahowald, Ryan Cotterell, and Dami{\'a}n
  Blasi. 2021{\natexlab{a}}.
\newblock \href {https://www.aclweb.org/anthology/2021.naacl-main.350} {How
  (non-)optimal is the lexicon?}
\newblock In \emph{Proceedings of the 2021 Conference of the North American
  Chapter of the Association for Computational Linguistics: Human Language
  Technologies}, pages 4426--4438, Online. Association for Computational
  Linguistics.

\bibitem[{Pimentel et~al.(2020)Pimentel, Roark, and
  Cotterell}]{pimentel-etal-2020-phonotactics}
Tiago Pimentel, Brian Roark, and Ryan Cotterell. 2020.
\newblock \href {https://doi.org/10.1162/tacl\_a\_00296} {Phonotactic
  complexity and its trade-offs}.
\newblock \emph{Transactions of the Association for Computational Linguistics},
  8:1--18.

\bibitem[{Pimentel et~al.(2021{\natexlab{b}})Pimentel, Roark, Wichmann,
  Cotterell, and Blasi}]{pimentel-etal-2021-finding}
Tiago Pimentel, Brian Roark, S{\o}ren Wichmann, Ryan Cotterell, and Dami{\'a}n
  Blasi. 2021{\natexlab{b}}.
\newblock \href {https://www.aclweb.org/anthology/2021.naacl-main.349} {Finding
  concept-specific biases in form{--}meaning associations}.
\newblock In \emph{Proceedings of the 2021 Conference of the North American
  Chapter of the Association for Computational Linguistics: Human Language
  Technologies}, pages 4416--4425, Online. Association for Computational
  Linguistics.

\bibitem[{Pitman and Yor(1997)}]{Pitman1997}
Jim Pitman and Marc Yor. 1997.
\newblock \href {https://doi.org/10.1214/aop/1024404422} {The two-parameter
  {P}oisson--{D}irichlet distribution derived from a stable subordinator}.
\newblock \emph{Annals of Probability}, 25(2):855--900.

\bibitem[{Takahashi and Tanaka-Ishii(2017)}]{takahashi2017neural}
Shuntaro Takahashi and Kumiko Tanaka-Ishii. 2017.
\newblock \href {https://doi.org/10.1371/journal.pone.0189326} {Do neural nets
  learn statistical laws behind natural language?}
\newblock \emph{{PLOS} {ONE}}, 12(12):1--17.

\bibitem[{Wei and Tanner(1990)}]{Wei1990}
Greg~C.G. Wei and Martin~A. Tanner. 1990.
\newblock \href {https://doi.org/10.2307/2290005} {A {M}onte {C}arlo
  implementation of the {EM} algorithm and the poor man's data augmentation
  algorithms}.
\newblock \emph{Journal of the American Statistical Association},
  85(411):699--704.

\bibitem[{Williams(1986)}]{williams1986origins}
Joseph~M. Williams. 1986.
\newblock \href
  {https://www.simonandschuster.com/books/Origins-of-the-English-Language/Joseph-M-Williams/9780029344705}
  {\emph{Origins of the {E}nglish language}}.
\newblock Simon and Schuster.

\bibitem[{Zipf(1935)}]{zipf1935psycho}
George~Kingsley Zipf. 1935.
\newblock \href {https://doi.org/10.4324/9781315009421} {\emph{The
  Psycho-biology of Language: {A}n Introduction to Dynamic Philology}}.
\newblock Houghton Mifflin.

\end{thebibliography}
\bibliographystyle{acl_natbib}

\vfill\null
\pagebreak

\newpage 
\appendix

\section{Tuned hyperparameres for the two-stage model}

\begin{table}[H]
    \centering
    \begin{tabular}{lrr}
        \toprule
         Language & a & b  \\
         \midrule
        English & 0.33 & 3,000 \\
        Finnish & 0.36 & 90,000 \\
        Hebrew & 0.40 & 55,000 \\
        Indonesian & 0.48 & 180,000\\
        Tamil & 0.70 & 37,000 \\
        Turkish & 0.33 & 95,000 \\
        Yoruba & 0.08 & 156,000 \\  \bottomrule
    \end{tabular}
    \caption{The optimized values of $a$ and $b$ for the analyzed languages.}
    \label{tab:tuned_hyperparams}
\end{table}

\section{Hyperparameter Search}

The same hyperparameters are used for both our baseline LSTMs and the generator. 
We use 3 layers, where embedding size is 128, hidden size is 512, and dropout probability is $0.33$.
Training the two-stage model takes a considerable amount of time (see \cref{tab:training_times}). We are thus not capable of doing exhaustive hyperparameter tuning. Random search \cite{Bergstra2012} is used in tuning the values for $a$ and $b$, where we run five training procedures considering ranges $a \in [0, 1)$, and $b \in [100, 200{,}000)$.
We tune the hyperparameters for each language by minimizing the model's cross-entropy on the development set, training them on a subset of the training data with only $100{,}000$ tokens.
The found optimal values of $a$ and $b$ are rounded to two decimal places and the thousands respectively.
Our two-stage model is trained for five iterations of expectation--maximization.

\section{Training time with the two-stage model for each language}

\begin{table}[H]
    \centering
    \begin{tabular}{lr}
        \toprule
         Language & Minutes\\
         \midrule
        English & 164  \\
        Finnish & 170 \\
        Hebrew & 175 \\
        Indonesian & 173\\
        Tamil & 166 \\
        Turkish & 174 \\
        Yoruba & 56 \\  \bottomrule
    \end{tabular}
    \caption{The training times for the two-stage model in each language. These times were obtained with a single NVIDIA Tesla P100 GPU.}
    \label{tab:training_times}
\end{table}

\section{The development set cross-entropies on the unigram distribution}

\begin{table}[H]
    \centering
\resizebox{\columnwidth}{!}{%
    \begin{tabular}{lcccc} \toprule
        & \multicolumn{4}{c}{Evaluated Model} \\ \cmidrule(lr){2-5}
        Language & Type & Token & Two-stage & Generator\\
        \midrule
        English & 12.50 & \phantom{0}9.04 & \phantom{0}\textbf{8.34} & \phantom{0}9.18 \\
        Finnish & 15.08 & 12.94 & \textbf{11.88} & 13.05  \\
        Hebrew & 12.62 & 10.71 & \textbf{10.20} & 10.78 \\
        Indonesian & 13.27 & 10.30 & \phantom{0}\textbf{9.53} & 10.42 \\
        Tamil & 14.22 & 12.65 & \textbf{11.76} & 12.75 \\
        Turkish & 13.99 & 11.74 & \textbf{10.83} & 12.92 \\
        Yoruba & 11.10 & 10.00 & \phantom{0}\textbf{9.13} & \phantom{0}9.83 \\  \bottomrule
    \end{tabular}
}
    \caption{Development set cross-entropy for the baseline models as well as our two-stage model evaluated on the \textbf{unigram distribution}.}
    \label{tab:token_dev_cross_entropies}
\end{table}

\section{Inference} \label{sec:inference}

Unfortunately, there is no closed form solution for inferring the parameters of our two-stage model.
In order to obtain a sample of cluster assignments and train the generator to match their labels, we estimate the parameters of both the generator and the adaptor concurrently, freezing one's parameters while training the other.
We use a regime corresponding to the Monte Carlo Expectation-maximization (EM) algorithm to train the model \cite{Wei1990}, which can be found in \cref{alg:general_training}.
In the E-step, the function \textsc{GibbsSampler} returns the cluster assignments $\bz$ and the dampened word dataset $\bl$ obtained via Gibbs sampling from the PYP.
We then use this dampened dataset to train the generator in the M-step.

\algrenewcommand{\algorithmiccomment}[1]{\hskip1em$\rightarrow$ \footnotesize#1 \normalsize}
\begin{algorithm}[H]
\begin{algorithmic}[1]
\For{$i$ \textbf{ in } $\textsc{range}(\text{\# Epochs})$} 
\State // E-Step
\State $\bz, \bl \sim \textsc{GibbsSampler}(a, b, \pgen, \{\bw_n\}_{1}^{N})$
\State // M-Step
\For{$t=1$ \textbf{ up to } $T$}
\State $\vphi \gets \eta_t\, \sum_{k=1}^{|\bl|} \nabla_{\vphi} \log \pgen(\bl_k \mid \vphi)$
\EndFor
\EndFor 
\end{algorithmic}
\caption{Training the two-stage model}
\label{alg:general_training}
\end{algorithm}

\subsection{Gibbs Sampler For Cluster Assignments}

The Pitman--Yor process does not have a well-defined posterior probability. 
Nonetheless, we can use Gibbs sampling to obtain a sample from this posterior distribution over cluster assignments defined by the two-stage model.\footnote{This is possible due to the exchangeability of the cluster assignments.}
We build our sampler after the morphological sampler presented by \newcite{Goldwater2011}.

\begin{figure*}
    \begin{align}  \label{eq:table_assignment_prob}
        p(\cluster \mid \otherclusters, \token) \propto p(\cluster, \token \mid \otherclusters) \propto
        \begin{cases}
            (c_{<n}^{(\cluster)}-a) \cdot \mathbbm{1}\{\token = \bl_{\cluster}\}\quad & 1 \leq \cluster \leq K_{<n} \\
            (a \cdot K_{<n} + b) \cdot \pgen(\token) \quad & \cluster = K_{<n} + 1 \\
        \end{cases} %
    \end{align}
    \caption{The probability of assigning token $\token$ to cluster $\cluster$ in the two-stage model given all other cluster assignments $\otherclusters$.}
    \label{fig:table_assignment_prob}
\end{figure*}

Gibbs sampling is a Markov Chain Monte Carlo (MCMC) method which approximates the posterior of a multivariate distribution. 
It iteratively samples from the conditional distribution of a variable, given the values of the other dimensions \cite{Neal1993}. 
We use the conditional distribution defined in \cref{eq:table_assignment_prob} (presented in \cref{fig:table_assignment_prob}) in the Gibbs sampler
where we know the word form $\token$ of token $n$---since it is observable in the corpus--- and where the values for all other cluster assignments are fixed.
Note that, according to \cref{eq:table_assignment_prob}, we only assign word tokens to clusters with the same form or create a new cluster---and when a new one is created, its word form is assigned to $\token$.
As such, each cluster contains a single shared word form.
For each adaptor training iteration, we run the Gibbs sampler for six epochs, and choose the cluster assignments that have the best performance on a development set. 
Furthermore, we persist the adaptor state across iterations, warm starting the Gibbs sampler with the cluster assignments of the previous iteration.

\subsection{Training the generator}

In order to train the generator on word form data with more balanced frequency distributions, a new training set is dynamically created. 
In this dataset, each token appears as many times as it has been assigned as a cluster label, noted with $\bl$ in \cref{alg:general_training}.\footnote{We hotstart the generator model by training it on a type-level dataset before the first adaptor training iteration.}
A regime similar to using the inverse-power transformed counts of the tokens in the corpus \cite{Goldwater2011}.

This new training set allows us to train the generator in an interpolation between a purely type- or token-based dataset; this interpolation can be controlled through its parameters $a$ and $b$.
Setting the values of $a$ and $b$ to zero will cause the model to favor existing clusters to creating new ones, resulting in assigning every token with the same form to a single cluster.
In this case, the generator parameters would be estimated using the equivalent of a \emph{type} corpus.
Similarly, when $a$ approaches one, or in the limit of $b \rightarrow \infty$, less tokens will be assigned per cluster and the number of single token clusters grows.
This is effectively equivalent to training the generator using \emph{tokens}. 
Consequently, non-extreme value of $a$ and $b$ are a middle ground.

We train the character-level LSTM used as our generator with stochastic gradient descent using a cross-entropy loss function.
This model is trained with early stopping; it is evaluated every 200 batches, and training stops when the development set loss has increased for 5 consecutive epochs.

\subsection{Training Optimizations}
The na\"{i}ve implementation of the Gibbs sampler for table assignments quickly becomes computationally expensive in practice.
Consequently, we use the optimized algorithm designed by \citet{blunsom-etal-2009-note} for the hierarchical Dirichlet process in our implementation, extending it to Pitman--Yor processes with the additional parameter $a$.

\section{Dataset} \label{sec:experimental_setup}

As noted in the main text, we use Wikipedia data in our experiments.
The amount of sentences used in our experiments is capped to one billion after shuffling them.
Additionally, we define an upper bound to the amount of tokens used in each experiment. 
In case the training data exceed this limit, we construct a corpus by re-sampling (with replacement) the desired number of tokens using the corpus frequencies calculated from the original training corpus.
The number of types and tokens used in training and evaluation are presented in \cref{tab:dataset_sizes}.
Noise in the Wikipedia data is somewhat reduced by hand-defining an alphabet for each language, and removing any sentence which includes words with invalid graphemes in it.%
\footnote{We define the alphabets using the languages' Wikipedia articles and the following website: \url{https://r12a.github.io/app-charuse/}.}

\vfill
\pagebreak

\begin{table}[H]
    \centering
\resizebox{\columnwidth}{!}{%
    \begin{tabular}{lrrrr} \toprule
        & \multicolumn{2}{c}{Train} & \multicolumn{2}{c}{Test} \\
        \cmidrule(r){2-3} \cmidrule(r){4-5}
         & \# Types & \# Tokens & \# Types & \# Tokens \\
        \midrule
        English & 76,589 & $10^6$ & 67,148 & 759,412 \\
        Finnish & 208,498 & $10^6$ & 108,020 & 332,220 \\
        Hebrew & 131,288 & $10^6$ & 105,550 & 619,685 \\
        Indonesian & 102,739 & $10^6$ & 72,250 & 507,848\\
        Tamil & 206,512 & $10^6$ & 116,165 & 388,257 \\
        Turkish & 154,185 & $10^6$ & 85,074 & 331,072\\
        Yoruba & 97,097 & 329,093 & 12,117 & 41,055\\  \bottomrule
    \end{tabular}
}
    \caption{The amount of tokens and types used in both training and testing for the analyzed languages.}
    \label{tab:dataset_sizes}
\end{table}

\end{document}